# Improved Definition of NonStandard Neutrosophic Logic

# and Introduction to Neutrosophic Hyperreals

# (Third version)

Prof. Florentin Smarandache, PhD, Postdoc
University of New Mexico
Mathematics Department
705 Gurley Ave., Gallup, NM 87301,USA

**Abstract**.

O the third version of this response-paper to Imamura's criticism, we recall that NonStandard Neutrosophic Logic was never used by neutrosophic community in no application, that the quarter of century old neutrosophic operators (1995-1998) criticized by Imamura were never utilized since they were improved shortly after but he omits to tell their development, and that in real world applications we need to convert/approximate the NonStandard Analysis hyperreals, monads and binads to tiny intervals with the desired accuracy – otherwise they would be inapplicable.

We point out several errors and false statements by Imamura [21] with respect to the inf/sup of nonstandard subsets, also Imamura's "rigorous definition of neutrosophic logic" is wrong and the same for his definition of nonstandard unit interval, and we prove that there is not a total order on the set of hyperreals (because of the newly introduced Neutrosophic Hyperreals that are indeterminate), whence the transfer principle is questionable.

After his criticism, several response publications on theoretical nonstandard neutrosophics followed in the period 2018-2022. As such, I extended the NonStandard Analysis by adding the *left monad closed to the right*, *right monad closed to the left*, *pierced binad* (we introduced in 1998), and *unpierced binad* - all these in order to close the newly extended nonstandard space ($R^*$) under nonstandard addition, nonstandard subtraction, nonstandard multiplication, nonstandard division, and nonstandard power operations [23, 24].

## 1. Introduction

I recall my first two answers to Imamura's 7th Nov. 2018 critics [1] about the NonStandard Neutrosophic Logic [20] on 24 Nov 2018 (version 1) and 13 Feb 2019 (version 2), and I update them after Imamura has published a third version [21] on a journal without even citing my previous response papers, nor making any comments or critics to them, although the paper was uploaded to arXiv shortly after him and also online at my UNM [20]. I find it as dishonest.

Surely, he can recall over and over again the first neutrosophic connectives, but he has to tell the whole story: they were never used in no application, and they were improved several times starting with the American researcher Ashbacher's neutrosophic connectives in 2002 that not even Rivieccio in 2008 was aware about.

The only reason I have added the nonstandard form to neutrosophic logic (and similarly to neutrosophic set and probability) was in order to make a distinction between *Relative Truth* (which is truth in some Worlds, according to Leibniz) and *Absolute Truth* (which is truth in all possible Words, according to Leibniz as well) that occur in philosophy.
Another possible reason may be when the neutrosophic degrees of truth, indeterminacy, or falsehood are infinitesimally determined, for example: the right monad ($0.8^+$) means a value strictly bigger than 0.8 but infinitely closer to 0.8. And similarly, the left monad ($^-0.8$) means a value strictly smaller than 0.8 but infinitely closer to 0.8. While the binad ($^-0.8^+$) means a value different from 0.8 but infinitely closer (from the right-hand side, or left-hand side) to 0.8. But they do not exist in our real world (the real set R), only in the hyperreal set R*, so we need to *convert / approximate* these hyperreal sets by tiny real intervals with the desired accuracy ( $\varepsilon$ ), such as: $(0.8, 0.8+\varepsilon)$ , $(0.8-\varepsilon, 0.8)$, or $(0.8-\varepsilon, 0.8) \cup (0.8, 0.8+\varepsilon)$ respectively [24].

Since the beginning of the neutrosophic field, many things have been developed and evolved, where better definitions, operators, descriptions, and applications of the neutrosophic logic have been defined. The same way happens in any scientific field: starting from some initial definitions and operations the community improves them little by little. The reader should check the last development of the neutroosphics - there are thousands of papers, books, and conference presentations online, check for example: http://fs.unm.edu/neutrosophy.htm. It is not fear to keep recalling the old definitions and operators since they have been improved in the meantime. The last development of the field should be revealed, not omitted.

The general definition of the neutrosophic set used in the last years.
Let *U* be a universe and a set *S* included in *U*. Then each element $x \in S$ , denoted as *x(T(x), I(x), F(x)),* has a degree of membership/truth T(x) with respect to *S*, degree of indeterminate-membership *I(x),* and degree of nonmembership *F(x),* where *T(x), I(x), F(x)* are real subsets of *[0, 1].*
I was more prudent when I presented the sum of single valued standard neutrosophic components, saying:

*Let T, I, F be single valued numbers, T, I, F* $\in$ *[0, 1], such that* $0 \leq T + I + F \leq 3$.

A friend alerted me: *"If T, I, F are numbers in [0, 1], of course their sum is between 0 and 3."*
"Yes, I responded, I afford this tautology, because if I did not mention that the sum is up to *3*, readers would take for granted that the sum *T + I + F* is bounded by *1*, since that is in all logics and in probability!"

Similarly, for the Neutrosophic Logic, but instead of elements we have propositions (in the propositional logic).

## 2. Errors in Imamura's paper [21]:

2.1. Imamura's assertation, referring to the Neutrosophic components *T, I, F* as subsets, that: "Subsets of $]^-0, 1^+[$" may have neither infima nor suprema" is false.

Counter-Examples of subsets that have both infima and suprema:
Let denote the nonstandard unit interval $U = ]^-0, 1^+[$.
Let $M = ]0.2^+, {}^-0.3[$, which is a subset of $U$, then
$inf(M) = 0.2$, $sup(M) = 0.3$.

In general, for any real numbers $a$ and $b$, such that $0 \leq a < b \leq 1$, one has
the corresponding nonstandard subset $S = ]a^+, {}^-b[$ included in $U$, that has both: $inf(S) = a$, $sup(S) = b$.
As a particular and interesting case, one has: $]0^+, {}^-1[ \subset ]^-0, 1^+[$.
Even more general, for any finite real numbers $a, b \in R$, $a < b$, the nonstandard subset $S = ]a^+, {}^-b[$ included in $R^*$, has both: $inf(S) = a$, $sup(S) = b$.

*2.2.* Imamura's "rigorous definition of neutrosophic logic" is wrong.

Let $\boldsymbol{K}$ be a nonarchimedean ordered field.
He defined, for $x, y \in K$, $x$ and $y$ are said to be <u>infinitely close</u> (denoted by $x \approx y$) if $x$-$y$ is infinitesimal. Then <u>$x$ is roughly smaller than $y$</u> (denoted as $x \underset{\approx}{<} y$) if $x < y$ or $x \approx y$.

This is wrong. See the below Counter-Examples.
Let $\varepsilon > 0$ be a positive infinitesimal, also $x = 5 + \varepsilon$ and $y = 5 - \varepsilon$ be hyperreals.
Of course, $x \in (5^+)$, right monad of $5$, and $y \in ({}^-5)$, left monad of $5$.
$5 + \varepsilon$ is infinitely closer to 5, but above (strictly greater than) 5;
while $5 - \varepsilon$ is infinitely closer to 5, but below (strictly smaller than) 5.
Then $x - y = 2\varepsilon$, which is infinitesimal,
And, because $x$ *is infinitely close to* $y$ ($x \approx y$), one has that $x$ is roughly smaller than $y$ (or $x \underset{\approx}{<} y$), according to Imamura's definition.
But this is false, since for $\varepsilon > 0$ clearly $5 + \varepsilon > 5 > 5 - \varepsilon$, whence $x > y$.
Therefore, x is not roughly smaller than y, but the opposite.

General Contra-Examples:
Let $\varepsilon > 0$ be a positive infinitesimal, and the real number $a \in R$.
Then for $x = a + \varepsilon$ and $y = a - \varepsilon$ we get the same wrong result $x < y$, according to Imamura.
Further on, for $x = a + \varepsilon$ and $y = a$, one gets the wrong result $x < y$.
And similarly, for $x = a$ and $y = a - \varepsilon$, one gets the wrong result $x < y$.

2.3. There exists no order between $a$ and ${}^-a^+$ in $R^*$.

Let $a \in R$ be a real number, and $\varepsilon$ be a positive or negative (we do not know exactly) infinitesimal.

Then $y = {}^-a^+$ is a hyperreal number of the form $y = a + \varepsilon$, where $\varepsilon$ may be positive or negative infinitesimal.

Let $({}^-a^+)$ be the left-right binad [5] of $a$, defined as:

$({}^-a^+) = \{a \pm \varepsilon,$ where $\varepsilon$ is a positive infinitesimal$\}$.

Of course, ${}^-a^+ \in ({}^-a^+)$.

The transfer principle [21] states that $R^*$ has the same first order properties as $R$.

But $R^*$ has only a partial order, since there is no order between $a$ and $^-a^+$ in $R^*$, while $R$ has a total order.

On has $\overset{-0}{a} \leq_{NSA} \overset{-0+}{a} \leq_{NSA} \overset{0+}{a}$, then $\overset{-0}{a} \leq_{NSA} a \leq_{NSA} \overset{0+}{a}$, whence $\overset{-0}{a} \leq_{NSA} \overset{0+}{a}$.

But, similar problems of non-order relationships are between $\overset{-0+}{a}$, $\overset{-0}{a}$, $\overset{0+}{a}$, and $^-a^+$.

Hence, the Transfer Principle from $R$ to $R^*$ is questionable…

## 3. Uselessness of Nonstandard Analysis in Neutrosophic Logic, Set, Probability. Statistics, et al.

Imamura's discussion [1] on the definition of neutrosphic logic is welcome, but it is useless, since from all neutrosophic papers and books published, from all conference presentations, and from all MSc and PhD theses defended around the world, etc. (more than two thousands) in the last two decades since the first neutrosophic research started *(1998-2022)*, and from thousands of neutrosophic researchers, not even a single one ever used the nonstandard form of neutrosophic logic, set, or probability and statistics in no occasion (extended researches or applications).

All researchers, with no exception, have used the *Standard Neutrosophic Set and Logic* [so no stance whatsoever of *Nonstandard Neutrosophic Set and Logic*], where the neutrosophic components *T, I, F* are real subsets of the standard unit interval *[0, 1].*

People don't even write "standard" since it is understood, because nonstandard was never used in no applications - it is unusable in real applications.

Even more, for simplifying the calculations, the majority of researchers have utilized the *Single-Valued Neutrosophic Set and Logic* {when *T, I, F* are single real numbers from *[0, 1]*}, on the second place was *Interval-Valued Neutrosophic Set and Logic* {when *T, I, F* are intervals included in *[0, 1]},* and on the third one the *Hesitant Neutrosophic Set and Logic* {when *T, I, F* were discrete finite subsets included in *[0, 1]*}.

In this direction, there have been published papers on single-valued "neutrosophic standard sets" [12, 13, 14], where the neutrosophic components are just *standard real numbers*, considering the particular case when $0 \leq T + I + F \leq 1$ (in the most general case $0 \leq T + I + F \leq 3$).

Actually, Imamura himself acknowledges on his paper [1], page 4, that:

> "neutrosophic logic does not depend on transfer, so the use of non-standard analysis is not essential for this logic, and can be eliminated from its definition".

Entire neutrosophic community has found out about this result and has ignored the nonstandard analysis from the beginning in the studies and applications of neutrosophic logic for two decades.

## 4. Applicability of Neutrosophic Logic et al. vs. Theoretical NonStandard Analysis

He wrote:

> "we do not discuss the theoretical significance or the applications of neutrosophic logic"

Why doesn't he discuss of the applications of neutrosophic logic? Because it has too many that brough its popularity among researchers [2], unlike the NonStandard Analysis that is a non-physical (idealistic, imaginary) object and it is hard to apply it in the real world.

Neutrosophic logic, set, measure, probability, statistics and so on were designed with the primordial goal of being applied in practical fields, such as:

Artificial Intelligence, Information Systems, Computer Science, Cybernetics, Theory Methods, Mathematical Algebraic Structures, Applied Mathematics, Automation, Control Systems, Big Data, Engineering, Electrical, Electronic, Philosophy, Social Science, Psychology, Biology, Biomedical, Engineering, Medical Informatics, Operational Research, Management Science, Imaging Science, Photographic Technology, Instruments, Instrumentation, Physics, Optics, Economics, Mechanics, Neurosciences, Radiology Nuclear, Medicine, Medical Imaging, Interdisciplinary Applications, Multidisciplinary Sciences etc. [2], while nonstandard analysis is mostly a pure mathematics.

Since 1990, when I emigrated from a political refugee camp in Turkey to America, working as a software engineer for Honeywell Inc., in Phoenix, Arizona State, I was advised by American coworkers to do theories that have *practical applications*, not pure-theories and abstractizations as "*art pour art*".

## 5. Theoretical Reason for the Nonstandard Form of Neutrosophic Logic

The only reason I have added the nonstandard form to neutrosophic logic (and similarly to neutrosophic set and probability) was in order to make a distinction between *Relative Truth* (which is truth in some Worlds, according to Leibniz) and *Absolute Truth* (which is truth in all possible Words, according to Leibniz as well) that occur in philosophy.

Another possible reason may be when the neutrosophic degrees of truth, indeterminacy, or falsehood are infinitesimally determined, for example a value infinitesimally bigger than *0.8 (*or $0.8^+$*)*, or infinitesimally smaller than *0.8 (*or $^-0.8$*)*. But these can easily be overcome by roughly using interval neutrosophic values and depending on the desired accuracy, for example (0.80, 0.81) and (0.79, 0.80) respectively.

I wanted to get the neutrosophic logic as general as possible [6], extending all previous logics (Boolean, fuzzy, intuitionistic fuzzy logic, intuitionistic logic, paraconsistent logic, dialethism), and to have it able to deal with all kinds of logical propositions (including paradoxes, nonsensical propositions, etc.).

That's why in 2013 I extended the Neutrosophic Logic to *Refined Neutrosophic Logic* [ from generalizations of *2-valued Boolean logic* to fuzzy logic, also from the *Kleene's and Lukasiewicz's* and *Bochvar's 3-symbol valued logics* or *Belnap's 4-symbol valued logic* to the most general *n-symbol* or *n-numerical* valued refined neutrosophic logic, for any integer $n \geq 1$ ], the largest ever so far, when some or all neutrosophic components *T, I, F* were respectively split/refined into neutrosophic subcomponents: $T_1, T_2, \ldots; I_1, I_2, \ldots; F_1, F_2, \ldots;$ which were deduced from our everyday life [3].

## 6. From Paradoxism movement to Neutrosophy – generalization of Dialectics

I started first from *Paradoxism* (that I founded in 1980's as a movement based on antitheses, antinomies, paradoxes, contradictions in literature, arts, and sciences), then I introduced the *Neutrosophy* (as generalization of Dialectics (studied by Hegel and Marx) and of Yin Yang (Ancient Chinese Philosophy), neutrosophy is a branch of philosophy studying the dynamics of triads, inspired from our everyday life, triads that have the form:

<*A*>, its opposite <*antiA*>, and their neutrals <*neutA*>,
where <*A*> is any item or entity [4].
(Of course, we take into consideration only those triads that make sense in our real and scientific world.)

The Relative Truth neutrosophic value was marked as $1$, while the Absolute Truth neutrosophic value was marked as $1^+$ (a tinny bigger than the Relative Truth's value):
$1^+ >_N 1$, where $>_N$ is a nonstandard inequality, meaning $1^+$ is nonstandardly bigger than $1$.
Similarly for Relative Falsehood / Indeterminacy (which falsehood / indeterminacy in some Worlds), and Absolute Falsehood / Indeterminacy (which is falsehood / indeterminacy in all possible worlds).

## 7. Introduction to Nonstandard Analysis [15, 16]

An *infinitesimal number* ($\varepsilon$) is a number $\varepsilon$ such that its absolute value $|\varepsilon| < 1/n$, for any non-null positive integer $n$. An infinitesimal is close to zero, and so small that it cannot be measured.

The infinitesimal is a number smaller, in absolute value, than anything positive nonzero.

Infinitesimals are used in calculus, but interpreted as tiny real numbers.

An *infinite number ($\omega$)* is a number greater than anything:

$$1 + 1 + 1 + \ldots + 1 \text{ (for any finite number terms)}$$

The infinites are reciprocals of infinitesimals.

The set of *hyperreals* (*non-standard reals*), denoted as $R^*$, is the extension of set of the real numbers, denoted as $R$, and it comprises the infinitesimals and the infinites, that may be represented on the *hyperreal number line*

$$1/\varepsilon = \omega/1.$$

The set of hyperreals satisfies the *transfer principle*, which states that the statements of first order in $R$ are valid in $R^*$ as well [according to the classical NonStandard Analysis].

A *monad* (*halo*) of an element $a \in R^*$, denoted by $\mu(a)$, is a subset of numbers infinitesimally close to $a$.

Let's denote by $R_+^*$ the set of positive nonzero hyperreal numbers.

### 7.1. **First Extension of NonStandard Analysis**

We consider the left monad and right monad; afterwards we recall the *pierced binad* (Smarandache [5]) introduced in 1998:

*Left Monad* {that we denote, for simplicity, by $(^{-}a)$} is defined as:

$$\mu(^{-}a) = (^{-}a) = \{a - x,\ x \in R_+^* \ /\ x \text{ is infinitesimal}\}.$$

*Right Monad* {that we denote, for simplicity, by $(a^+)$} is defined as:

$$\mu(a^+) = (a^+) = \{a + x,\ x \in R_+^* \ /\ x \text{ is infinitesimal}\}.$$

The *Pierced Binad* {that we denote, for simplicity, by $({}^{-}a^{+})$} is defined as:

$$\mu({}^{-}a^{+}) = ({}^{-}a^{+}) = \{a - x,\ x \in R_{+}^{*} \ /\ x \text{ is infinitesimal}\} \cup \{a + x,\ x \in R_{+}^{*} \ /\ x \text{ is infinitesimal}\}$$

$$= \{a - x,\ x \in R^{*} \ /\ x \text{ is positive or negative infinitesimal}\}.$$

### 7.2. Second Extension of Nonstandard Analysis [23]

For necessity of doing calculations that will be used in nonstandard neutrosophic logic in order to calculate the nonstandard neutrosophic logic operators (conjunction, disjunction, negation, implication, equivalence) and in order to have the Nonstandard Real MoBiNad Set closed under arithmetic operations, we extend now for the time: the left monad to the Left Monad Closed to the Right, the right monad to the Right Monad Closed to the Left; and the Pierced Binad to the Unpierced Binad, defined as follows (Smarandache, 2018-2019):

**Left Monad Closed to the Right**

$$\mu\begin{pmatrix} {}^{-0} \\ a \end{pmatrix} = \begin{pmatrix} {}^{-0} \\ a \end{pmatrix} = \{a - x \mid x = 0, \text{ or } x \in R_{+}^{*} \textit{ and x is infinitesimal}\} = \mu({}^{-}a) \cup \{a\}.$$

And by $x = \overset{-0}{a}$ we understand the **hyperreal** $x = a - \varepsilon$, or $x = a$, where $\varepsilon$ is a positive infinitesimal. So, $x$ is not clearly known, $x \in \{a - \varepsilon, a\}$.

**Right Monad Closed to the Left**

$$\mu\begin{pmatrix} {}^{0+} \\ a \end{pmatrix} = \begin{pmatrix} {}^{0+} \\ a \end{pmatrix} = \{a + x \mid x = 0, \text{ or } x \in R_{+}^{*} \textit{ and x is infinitesimal}\} = \mu(a^{+}) \cup \{a\}.$$

And by $x = \overset{-0}{a}$ we understand the **hyperreal** $x = a + \varepsilon$, or $x = a$, where $\varepsilon$ is a positive infinitesimal. So, $x$ is not clearly known, $x \in \{a + \varepsilon, a\}$.

**Unpierced Binad**

$$\mu\begin{pmatrix} {}^{-0+} \\ a \end{pmatrix} = \begin{pmatrix} {}^{-0+} \\ a \end{pmatrix} = \{a + x \mid x = 0, \text{or } x \in R^{*} \textit{ where x is a positive or negative infinitesimal}\} =$$

$$= \mu\left({}^{-}a\right) \cup \mu\left(a^{+}\right) \cup \{a\} = ({}^{-}a) \cup (a^{+}) \cup \{a\}.$$

And by $x = \overset{-0+}{a}$ we understand the **hyperreal** $x = a - \varepsilon$, or $x = a$, or $x = a + \varepsilon$, where $\varepsilon$ is a positive infinitesimal. So, $x$ is not clearly known, $x \in \{a - \varepsilon, a, a + \varepsilon\}$.

The left monad, left monad closed to the right, right monad, right monad closed to the left, the pierced binad, and the unpierced binad are subsets of $R^{*}$, while the above hyperreals are numbers from $R^{*}$.

Let's define a partial order on $R^{*}$.

## 8. Neutrosophic Strict Inequalities

We recall the neutrosophic strict inequality which is needed for the inequalities of nonstandard numbers.

Let $\alpha, \beta$ be elements in a partially ordered set $M$.

We have defined the *neutrosophic strict inequality*

$\alpha >_N \beta$

and read as

*"α is neutrosophically greater than β"*

if

*α in general is greater than β,*

or *α is approximately greater than β,*

or *subject to some indeterminacy* (unknown or unclear ordering relationship between α and β) *or subject to some contradiction (*situation when α is smaller than or equal to β*) α is greater than β.*

It means that in most of the cases, on the set *M*, *α is greater than β.*

And similarly for the opposite neutrosophic strict inequality $\alpha <_N \beta$.

## 9. Neutrosophic Equality

We have defined the *neutrosophic inequality*

$\alpha =_N \beta$

and read as

*"α is neutrosophically equal to β"*

if

*α in general is equal to β,*

or *α is approximately equal to β,*

or *subject to some indeterminacy* (unknown or unclear ordering relationship between α and β) *or subject to some contradiction (*situation when α is not equal to β*) α is equal to β.*

It means that in most of the cases, on the set *M*, *α is equal to β.*

## 10. Neutrosophic (Non-Strict) Inequalities

Combining the neutrosophic strict inequalities with neutrosophic equality, we get the $\geq_N$ and $\leq_N$ neutrosophic inequalities.

Let $\alpha, \beta$ be elements in a partially ordered set $M$.

The *neutrosophic (non-strict) inequality*

$\alpha \geq_N \beta$

and read as

"*α is neutrosophically greater than or equal to β*"

if

*α in general is greater than or equal to β,*

or *α is approximately greater than or equal to β,*

or *subject to some indeterminacy* (unknown or unclear ordering relationship between α and β) *or subject to some contradiction (*situation when α is smaller than β*) α is greater than or equal to β.*

It means that in most of the cases, on the set $M$, *α is greater than or equal to β.*

And similarly for the opposite neutrosophic (non-strict) inequality $\alpha \leq_N \beta$.

## 11. Neutrosophically Ordered Set

Let $M$ be a set. $(M, <_N)$ is called a neutrosophically ordered set if:
$\forall\ \alpha, \beta \in M$, one has: either $\alpha <_N \beta$, or $\alpha =_N \beta$, or $\alpha >_N \beta$.

## 12. Definition of Standard Part and Infinitesimal Part of a HyperReal Number

For each hyperreal (number) $h \in R^*$ one defines its <u>standard part</u>
$st(h)$ be the real (standard) part of h, $st(h) \in R$,
and its <u>infinitesimal part</u>, that may be positive $(+\varepsilon)$, or zero (0), or negative $(-\varepsilon)$, and any combination of two or three of them in the case of Neutrosophic Hyperreals that have alternative (indeterminate) values as seen below, denoted as $in(h) \in R^*$.
Then $h = st(h) + in(h)$.
Two hyperreal numbers $h_1$ and $h_2$ are equal, if:
$st(h_1) = st(h_2)$ *and* $in(h_1) = in(h_2)$.

**Examples**:
Let $\varepsilon$ be a positive infinitesimal, and the hyperreal numbers:
$h_1 = 4 - \varepsilon \in ({}^-4)$
$h_2 = 4 + 0 \overset{def\ \ 0}{=} 4 \in R$
$h_3 = 4 + \varepsilon \in (4^+)$

$h_4 = 4 - \{\varepsilon, \text{or } 0\}$ = {4- $\varepsilon$ , or 4-0} = {4- $\varepsilon$ , or 4} $\in \left( \overset{-0}{4} \right)$

$h_5 = 4 + \{0, \text{or } \ \varepsilon \ \}$ = {4+0, or 4+ $\varepsilon$ } = {4, or 4+ $\varepsilon$ } $\in \left( \overset{0+}{4} \right)$

$h_6 = 4 + \{-\varepsilon, \text{or } \ \varepsilon \ \}$ = {4- $\varepsilon$ , or 4+ $\varepsilon$ } $\in \left( \overset{-+}{4} \right)$

$h_7 = 4 + \{-\varepsilon, \text{or } 0, \text{ or } \ \varepsilon \ \}$ = {4- $\varepsilon$ , or 4+0, or 4+ $\varepsilon$ }= {4- $\varepsilon$ , or 4, or 4+ $\varepsilon$ } $\in \left( \overset{-0+}{4} \right)$

Then, their standard parts are all the same:
$st(h_1) = st(h_2) = ... = st(h_7) = 4$

While their infinitesimal parts are different:
$in(h_1) = -\varepsilon$
$in(h_2) = 0$
$in(h_3) = \varepsilon$

## 13. Neutrosophic Hyperreal Numbers

The below cases are indeterminate, as in neutrosophy, that's why they are called *Neutrosophic Hyperreals*, introduced now for the first time:
$in(h_4) = \{-\varepsilon$ , or $0\}$; one can also write that $in(h_4) \in \{-\varepsilon, 0\}$, because we are not sure if $in(h_4) = -\varepsilon$ , or $in(h_4) = 0$.
$in(h_5) = \{\varepsilon$ , or $0\}$; one can also write that $in(h_4) \in \{\varepsilon, 0\}$.
$in(h_6) = \{-\varepsilon, \text{ or } \ \varepsilon \ \}$, or $in(h_6) \in \{-\varepsilon, \varepsilon\}$.
$in(h_7) = \{-\varepsilon, \text{ or } 0, \text{ or } \ \varepsilon \ \}$, or $in(h_6) \in \{-\varepsilon, 0, \varepsilon\}$.

## 14. Nonstandard Partial Order of Hyperreals

Let $h_1$ and $h_2$ be hyperreal numbers. Then $h_1 <_N h_2$ if:

either $st(h_1) < st(h_2)$, or $st(h_1) = st(h_2)$ *and* $in(h_1) <_N in(h_2)$.

By $in(h_1)$ we understand all possible infinitesimals of $h_1$, and similarly for $in(h_2)$.

This makes a partial order on the set of hyperreals $R^*$, because of the Neutrosophic Hyperreals that have indeterminate infinitesimal parts and cannot always be ordered.

## 15. Appurtenance of a Hyperreal number to a Nonstandard Set.

We define for the first time the appurtenance of a hyperreal number ($h$) to a subset $S$ of $R^*$, denoted as $\in_N$ , or an approximate appurtenance (from a Neutrosophic point of view).
As seeing above, a hyperreal number may have one, two, or three infinitesimal parts - depending on its form.
Let's denote the standard part of $h$ by $st(h)$, and its infinitesimal part(s) be $in(h) = in(h)_1$, $in(h)_2$, and $in(h)_3$. We construct three corresponding hyperreal numbers:
$h_1 = st(h) + in(h)_1$

$h_2 = st(h) + in(h)_2$
$h_3 = st(h) + in(h)_3$
If all three $h_1, h_2, h_3 \in_N S$, then $h \in_N S$. If at least one does not belong to *S*, then $h \notin_N S$.
(In the case when *h* has only one or two possible infinitesimals, of course we take only them.)

The appurtenance of a hyperreal number to a nonstandatd set may be later extended if new forms of Neutrosophic Hyperreals are constructed in the meantime.

## 15. Notations and Approximations

Approximation is required with a desired accuracy, since the hyperreals, monads and binads do not exist in our real world. They are only very abstract concepts built in some imaginary math space.
That's why they must be approximated by real tiny sets.
As an example, let's assume that the truth-value (T) of a proposition (P), in the propositional logic, is the hyperreal T(P) = $0.7^+$ that means, in nonstandard analysis, according to Imamura [22]:

> "The interpretation of T(P) = $0.7^+$ (right monad of 0.7 in your terminology):
> 1. the truth value of P is strictly greater than and infinitely close to 0.7 (but its precise value is unknown);
> 2. the truth value of P can be strictly greater than and infinitely close to 0.7;
> 3. the truth value of P takes all hyperreals strictly greater than and infinitely close to 0.7 simultaneously."

We prove by reductio ad absurdum that such a number does not exist in our real world. Let assume that $0.7^+ = w$. Then $w > 0.7$, but on the set of continuous real numbers, in the interval (0.7, w] there exists a number v such that $0.7 < v < w$, therefore v is closer to 0.7 than w, and thus w is not infinitely close to 0.7. Contradiction. Even Imamura acknowledges about $0.7^+$ that "its value is unknown".

And because they do not exist in our real world, we need to approximate/convert them with a given accuracy to the real world, therefore, instead of $0.7^+$ we may take for example the tiony interval (0.7, 0.7001) with four decimals, or (0.7, 0.7000001), etc.

In the same way one can prove that, for any real number $a \in R$, its left monad, left monad closed to the right, right monad, right monad closed to the left, pierced binad, and unpierced binad do not exist in our real world. They are just abstract concepts available in abstract/imaginary math spaces.

### 17. Nonstandard Unit Interval.

Imamura cites my work:

"by "$_{-}a$" one signifies a monad, i.e., a set of hyper-real numbers in non-standard analysis: $(_{-}a) = \{ a - x \in R_{*} \mid x \text{ is infinitesimal} \}$, and similarly "$b_{+}$" is a hyper monad: $(b_{+}) = \{ b + x \in R_{*} \mid x \text{ is infinitesimal} \}$. ([5] p. 141; [6] p. 9)"

But these are inaccurate, because my exact definitions of monads, since my 1998 first world neutrosophic publication {see [5], page 9; and [6], pages 385 - 386}, were:

"$(^{-}a) = \{ a - x: x \in R_{+}^{*} \mid x \text{ is infinitesimal} \}$, and similarly "b+" is a hyper monad: $(b^{+}) = \{ b + x: x \in R_{+}^{*} \mid x \text{ is infinitesimal} \}$"

Imamura says that:

"The correct definitions are the following:
$(_{-}a) = \{ a - x \in R_{*} \mid x \text{ is } \textit{positive} \text{ infinitesimal} \}$,
$(b_{+}) = \{ b + x \in R_{*} \mid x \text{ is } \textit{positive} \text{ infinitesimal} \}$."

I did not have a chance to see how my article was printed in *Proceedings of the 3rd Conference of the European Society for Fuzzy Logic and Technology* [7], that Imamura talks about, maybe there were some typos, but Imamura can check the *Multiple Valued Logic / An International Journal* [6], published in England in 2002 (ahead of the European Conference from 2003, that Imamura cites) by the prestigious Taylor & Francis Group Publishers, and clearly one sees that it is: $\boldsymbol{R}_{+}^{*}$ (so, $x$ is a *positive* infinitesimal into the above formulas), therefore there is no error.

Then Imamura continues:

"Ambiguity of the definition of the nonstandard unit interval. Smaran- dache did not give any explicit definition of the notation $]^{-}0, 1^{+}[$ in [5] (or the notation ⊩$^{-}0, 1^{+}$⫣ in [6]). He only said:
Then, we call $]^{-}0, 1^{+}[$ a non-standard unit interval. Obviously, 0 and 1, and analogously non-standard numbers infinitely small but less than 0 or infinitely small but greater than 1, belong to the non-standard unit interval. ([5] p. 141; [6] p. 9)."

Concerning the notations I used for the nonstandard intervals, such as ⊩ ⫣ or ] [, it was imperative to employ notations that are different from the classical *[ ]* or *( )* intervals, since the extremes of the nonstandard unit interval were unclear, vague with respect to the real set.
I thought it was easily understood that:

$$]^{-}0, 1^{+}[ \; = (^{-}0) \cup [0, 1] \cup (1^{+}).$$

Or, using the previous neutrosophic inequalities, we may write:

$$]^{-}0, 1^{+}[ \; = \{x \in R^{*}, \; ^{-}0 \leq_{N} x \leq_{N} 1^{+}\}.$$

Imamura says that:

> "Here $^{-}0$ and $1^{+}$ are particular real numbers defined in the previous paragraph: $^{-}0 = 0-\varepsilon$ and $1^{+} = 1+\varepsilon$, where ε is a fixed non-negative infinitesimal."

This is untrue, I never said that "ε is a *fixed* non-negative infinitesimal", ε was not fixed, I said that for any real numbers *a* and *b* {see again [5], page 9; and [6], pages 385 - 386}:

$$\text{"}(^{-}a) = \{\, a - x \colon x \in R_{+}^{*} \mid x \text{ is infinitesimal } \},$$
$$(b^{+}) = \{\, b + x \colon x \in R_{+}^{*} \mid x \text{ is infinitesimal } \}\text{"}.$$

Therefore, once we replace $a = 0$ and $b = 1$, we get:

$$(^{-}0) = \{\, 0 - x \colon x \in R_{+}^{*} \mid x \text{ is infinitesimal } \},$$
$$(1^{+}) = \{\, 1 + x \colon x \in R_{+}^{*} \mid x \text{ is infinitesimal } \}.$$

Thinking out of box, inspired from the real world, was the first intent, i.e. allowing neutrosophic components (truth / indeterminacy / falsehood) values be outside of the classical (standard) unit real interval *[0, 1]* used in all previous (Boolean, multi-valued etc.) logics if needed in applications, so neutrosophic component values $< 0$ and $> 1$ had to occurs due to the Relative / Absolute stuff, with:

$$^{-}0 <_N 0 \quad \text{and} \quad 1^{+} >_N 1.$$

Later on, in 2007, I found plenty of cases and real applications in Standard Neutrosophic Logic and Set (therefore, not using the Nonstandard Neutrosophic Logic and Set), and it was thus possible the extension of the neutrosophic set *to Neutrosophic Overset (when some neutrosophic component is > 1), and to Neutrosophic Underset (when some neutrosophic component is < 0), and to Neutrosophic Offset (when some neutrosophic components are off the interval [0, 1], i.e. some neutrosophic component > 1 and some neutrosophic component < 0). Then, similar extensions to respectively Neutrosophic Over/Under/Off Logic, Measure, Probability, Statistics etc.* [8, 17, 18, 19], extending the unit interval [0, 1] to

$[\Psi, \Omega]$, with $\Psi \le 0 < 1 \le \Omega$,

where $\Psi, \Omega$ are standard real numbers.

Imamura says, regarding the definition of neutrosophic logic that:

> "In this logic, each proposition takes a value of the form (T, I, F), where T, I, F are subsets of the nonstandard unit interval $]^{-}0, 1^{+}[$ and represent all possible values of Truthness, Indeterminacy and Falsity of the proposition, respectively."

Unfortunately, this is not exactly how I defined it.

In my first book {see [5], p. 12; or [6] pp. 386 – 387} it is stated:

> "Let T, I, F be real standard or non-standard subsets of ]-0, 1+["

meaning that T, I, F may also be "real standard" not only real non-standard.

In *The Free Online Dictionary of Computing*, 1999-07-29, edited by Denis Howe from England, it is written:

> Neutrosophic Logic:
> <*logic*> (Or "Smarandache logic") A generalization of fuzzy logic based on Neutrosophy. A proposition is t true, i indeterminate, and f false, where t, i, and f are real values from the ranges T, I, F, with no restriction on T, I, F, or the sum n = t + i + f. Neutrosophic logic thus generalizes:
> - intuitionistic logic, which supports incomplete theories (for 0 < n < 100, 0 ≤ t,i,f ≤ 100);
> - fuzzy logic (for n = 100 and i = 0, and 0 ≤ t,i,f ≤ 100);
> - Boolean logic (for n=100 and i = 0, with t,f either 0 or 100);
> - multi-valued logic (for 0 ≤ t,i,f ≤ 100);
> - paraconsistent logic (for n > 100, with both t,f < 100);
> - dialetheism, which says that some contradictions are true (for t = f = 100 and i = 0; some paradoxes can be denoted this way).
>
> Compared with all other logics, neutrosophic logic introduces a percentage of "indeterminacy" - due to unexpected parameters hidden in some propositions. It also allows each component t,i,f to "boil over" 100 or "freeze" under 0. For example, in some tautologies t > 100, called "overtrue".
> *Home*.
> ["Neutrosophy / Neutrosophic probability, set, and logic", F. Smarandache, American Research Press, 1998].

As Denis Howe said in 1999, the neutrosophic components *t, i, f* are "*real values* from the ranges *T, I, F*", not nonstandard values or nonstandard intervals. And this was because nonstandard ones were not important for the neutrosophic logic (the Relative/Absolute plaid no role in technological and scientific applications and future theories).

## 18. Formal Notations:

In my first version of the paper, I used informal notations. Let's see them improved.

Hyperreal Numbers:

$${}^{-}a = \overset{-}{a} = a - \varepsilon$$

$\overset{0}{a} = a + 0$, which coincides with the real number *a.*

$$a^{+} = \overset{+}{a} = a + \varepsilon$$

Neutrosophic Hyperreal Numbers (that are indeterminate, alternative):

$\overset{-0}{a} = a - \varepsilon$, or $a + 0$

$\overset{+0}{a} = a+\varepsilon$, or $a + 0$

$\overset{-+}{a} = a-\varepsilon$, or $a+\varepsilon$

$\overset{-0+}{a} = a-\varepsilon$, or $a + 0$, or $a+\varepsilon$

For the monads and binads one just adds the parentheses:

Monad Sets: $a = \left(\overset{0}{a}\right), (^{-}a) = \left(\overset{-}{a}\right), (a^{+}) = \left(\overset{+}{a}\right)$

Binad Sets: $\left(\overset{-0}{a}\right), \left(\overset{0+}{a}\right), \left(\overset{-+}{a}\right), \left(\overset{-0+}{a}\right)$

**19. Improved Definition of NonStandard Unit Interval**

$$]^{-}0, 1^{+}[ \equiv ]\overset{-}{0}, \overset{+}{1}[ = \{a \in R^{*}, 0 \le st(a) \le 1\} =$$

$$= \{\overset{-}{a}, \overset{-0}{a}, \overset{0}{a}, \overset{0+}{a}, \overset{-+}{a}, \overset{-0+}{a}, \overset{+}{a}, a \in [0,1]\}.$$

It is not necessarily to set any restriction on *in(a)* in this case, since $\overset{-}{a}$ is the smallest hyperreal, while $\overset{+}{a}$ is the greatest hyperreal in the set of seven hyperreals listed above.

An example:

$]\overset{+}{0}, \overset{-}{1}[ \subset ]0,1[ \subset ]\overset{-}{0}, \overset{+}{1}[$

Let $a \in R$ be a real number. Then there is no order between $a$ and $\overset{-+}{a}$, nor between $a$ and $\overset{-0+}{a}$.
Some nonstandard inequalities involving hyperreals:

$\overset{-}{a} <_{N} \overset{0}{a} <_{N} \overset{+}{a}$

$\overset{-0}{a} \le_{N} \overset{-+}{a} \le_{N} \overset{0+}{a} \le_{N} \overset{+}{a}$

$\overset{-}{a} \le_{N} \overset{-0}{a} \le_{N} \overset{-+}{a} \le_{N} \overset{-0+}{a}$

$\overset{-}{a} \le_{N} \overset{-+}{a} \le_{N} \overset{+}{a}$

Examples of Nonstandard Intervals:

$$]\overset{-}{a},\overset{0}{a}[=\{\overset{-}{a},\overset{0}{a},\overset{-0}{a}\}$$

$$]\overset{-}{a},\overset{+}{a}[=\{\overset{-}{a},\overset{0}{a},\overset{+}{a},\overset{-0}{a},\overset{0+}{a},\overset{-+}{a},\overset{-0+}{a}\}$$

## 20. The Logical Connectives ∧, ∨, →

Imamura's critics of my first definition of the neutrosophic operators is history for over a quarter of century ago. He is attacking my paper with "errors… errors" etc., but my first operators were not kind of errors, but less accurate approximations of the aggregation with respect to the falsity component (F), but not with respect to the truth (T) and indeterminacy (I) ones that were correct.

The representations of the monads and binads by tiny intervals were also approximations with a desired accuracy ($\varepsilon$), from a classical (real) point of view:

$(^{-}a) \cong (a-\varepsilon, a)$,

$(b^{+}) \cong (b, b+\varepsilon)$,

$(^{-}a^{+}) \cong (a-\varepsilon, a) \cup (b, b+\varepsilon)$.

All aggregations in fuzzy and fuzzy-extensions (that includes neutrosophic) logics and sets are *approximations* (not exact, as in classical logic), and they depend on each specific application and on the experts. Some experts/authors prefer ones, other authors prefer other operators.
It is NOT A UNIQUE operator of fuzzy/neutrosophic conjunction, as it is in classical logic, but a class of many neutrosophic operators, which is called neutrosophic t-norm; similarly for fuzzy/neutrosophic disjunction, called neutrosophic t-conorm, fuzzy/neutrosophic negation, fuzzy/neutrosophic implication, fuzzy/neutrosophic equivalence, etc.
All fuzzy, intuitionistic fuzzy, neutrosophic (and other fuzzy-extension) logic operators are *inferential approximations*, not written in stone. They are improved from application to application.

Let's denote:

$\wedge_F, \wedge_N, \wedge_P$ representing respectively the fuzzy conjunction, neutrosophic conjunction, and plithogenic conjunction; similarly

$\vee_F, \vee_N, \vee_P$ representing respectively the fuzzy disjunction, neutrosophic disjunction, and plithogenic disjunction, and

$\rightarrow_F, \rightarrow_N, \rightarrow_P$ representing respectively the fuzzy implication, neutrosophic implication, and plithogenic implication.

I agree that my beginning neutrosophic operators (when I applied the same *fuzzy t-norm*, or the same *fuzzy t-conorm*, to all neutrosophic components *T, I, F*) were less accurate than others developed later by the neutrosophic community researchers. This was pointed out since 2002 by Ashbacher [9] and confirmed in 2008 by Rivieccio [10] much ahead of Imamura [1] in 2018. They observed that if on $T_1$ and $T_2$ one applies a *fuzzy t-norm*, on their opposites $F_1$ and $F_2$ one needs to apply the *fuzzy t-conorm* (the opposite of fuzzy t-norm), and reciprocally.

About inferring $I_1$ and $I_2$, some researchers combined them in the same directions as $T_1$ and $T_2$.

Then:

$$(T_1, I_1, F_1) \wedge_N (T_2, I_2, F_2) = (T_1 \wedge_F T_2, I_1 \wedge_F I_2, F_1 \vee_F F_2),$$
$$(T_1, I_1, F_1) \vee_N (T_2, I_2, F_2) = (T_1 \vee_F T_2, I_1 \vee_F I_2, F_1 \wedge_F F_2),$$
$$(T_1, I_1, F_1) \rightarrow_N (T_2, I_2, F_2) = (F_1, I_1, T_1) \vee_N (T_2, I_2, F_2) = (F_1 \vee_F T_2, I_1 \vee_F I_2, T_1 \wedge_F F_2);$$

others combined $I_1$ and $I_2$ in the same direction as $F_1$ and $F_2$ (since both $I$ and $F$ are negatively qualitative neutrosophic components), the most used one:

$$(T_1, I_1, F_1) \wedge_N (T_2, I_2, F_2) = (T_1 \wedge_F T_2, I_1 \vee_F I_2, F_1 \vee_F F_2),$$
$$(T_1, I_1, F_1) \vee_N (T_2, I_2, F_2) = (T_1 \vee_F T_2, I_1 \wedge_F I_2, F_1 \wedge_F F_2),$$
$$(T_1, I_1, F_1) \rightarrow_N (T_2, I_2, F_2) = (F_1, I_1, T_1) \vee_N (T_2, I_2, F_2) = (F_1 \vee_F T_2, I_1 \wedge_F I_2, T_1 \wedge_F F_2).$$

Now, applying the neutrosophic conjunction suggested by Imamura:

> "This causes some counterintuitive phenomena. Let A be a (true) proposition with value ({ 1 } , { 0 } , { 0 }) and let B be a (false) proposition with value ({ 0 } , { 0 } , { 1 }).
> Usually we expect that the falsity of the conjunction A ∧ B is { 1 }. However, its actual falsity is { 0 }."

we get:

$$(1, 0, 0) \wedge_N (0, 0, 1) = (0, 0, 1), \quad (50)$$

which is correct (so the falsity is *1*).

Even more, recently, in an extension of neutrosophic set to *plithogenic set* [11] (which is a set whose each element is characterized by many attribute values), the *degrees of contradiction c( , )* between the neutrosophic components *T, I, F* have been defined (in order to facilitate the design of the aggregation operators), as follows: $c(T, F) = 1$ (or *100%,* because they are totally opposite), $c(T, I) = c(F, I) = 0.5$ (or *50%,* because they are only half opposite), then:

$$(T_1, I_1, F_1) \wedge_P (T_2, I_2, F_2) = (T_1 \wedge_F T_2, 0.5(I_1 \wedge_F I_2) + 0.5(I_1 \vee_F I_2), F_1 \vee_F F_2),$$
$$(T_1, I_1, F_1) \vee_P (T_2, I_2, F_2) = (T_1 \vee_F T_2, 0.5(I_1 \vee_F I_2) + 0.5(I_1 \wedge_F I_2), F_1 \wedge_F F_2).$$
$$(T_1, I_1, F_1) \rightarrow_N (T_2, I_2, F_2) = (F_1, I_1, T_1) \vee_N (T_2, I_2, F_2)$$
$$= (F_1 \vee_F T_2, 0.5(I_1 \vee_F I_2) + 0.5(I_1 \wedge_F I_2), T_1 \wedge_F F_2).$$

**Conclusion**

We thank very much Dr. Takura Imamura for his interest and critics of *Nonstandard Neutrosophic Logic*, which eventually helped in improving it. {In the history of mathematics, critics on nonstandard analysis, in general, have been made by Paul Halmos, Errett Bishop, Alain Connes and others.} We hope we'll have more dialogues on the subject in the future.

We introduced for the first time the Neutrosophic Hyperreals (that have an indeterminate form).

We pointed out several errors and false statements by Imamura [21] with respect to the inf/sup of nonstandard subsets, also Imamura's "rigorous definition of neutrosophic logic" is wrong and the same for his definition of nonstandard unit interval, and we proved that there is not a total order on the set of hyperreals (because of the newly introduced Neutrosophic Hyperreals that are indeterminate) therefore the transfer principle is questionable.